\documentclass{article}

\usepackage[preprint]{neurips_2026} % Uncomment for pre-prints (e.g., arxiv)

\usepackage[utf8]{inputenc} % allow utf-8 input
\usepackage[T1]{fontenc}    % use 8-bit T1 fonts
\usepackage{hyperref}       % hyperlinks
\usepackage{url}            % simple URL typesetting
\usepackage{booktabs}       % professional-quality tables
\usepackage{amsfonts}       % blackboard math symbols
\usepackage{nicefrac}       % compact symbols for 1/2, etc.
\usepackage{microtype}      % microtypography
\usepackage{xcolor}         % colors

\usepackage{amsmath}
\usepackage{amsthm}
\usepackage{amssymb}
\usepackage{graphicx}
\usepackage{stmaryrd}
\usepackage{algorithm}
\usepackage{algpseudocode}
\usepackage{wrapfig}
\usepackage{subcaption}
\usepackage{enumitem}
\usepackage[bottom]{footmisc}

\def\mybigtimes{\mathop{\mathchoice{%display:
   \vcenter{\hbox to10bp{\vrule height13bp width0pt \pdfliteral{
   q 1 J .8 w 0 1 m 10 14 l S 0 14 m 10 1 l S Q
}\hss}}}{%text:
   \vcenter{\hbox to10bp{\kern1bp\vrule height10bp width0pt \pdfliteral{
   q 1 J .65 w 0 0 m 8 10 l S 0 10 m 8 0 l S Q
}\hss}}}{\times}{\times}%script, scriptscript not defined
}}

\title{Robust Instruction Compliance in Cooperative Multi-Agent Reinforcement Learning}

\author{%
  Wo Wei Lin\\
  Department of Computer Sciences\\
  Northeastern University \\
  United States\\
  \texttt{lin.wo@northeastern.edu} \\
  \And
  Ethan Rathbun\\
  Department of Computer Sciences\\
  Northeastern University \\
  United States\\
  \texttt{rathbun.e@northeastern.edu} \\
  \AND
  Enrico Marchesini \\
  Department of Computer Science\\
  University of Verona\\
  Italy\\
  \texttt{enrico.marchesini@univr.it} \\
  \And
  Xiang Zhi Tan \\
  Department of Computer Sciences\\
  Northeastern University \\
  United States\\
  \texttt{zhi.tan@northeastern.edu}
}

\begin{document}
\maketitle

\begin{abstract}
Multi-agent reinforcement learning (MARL) in real-world use cases may need to adapt to external natural language instructions that interrupt ongoing behavior and conflict with long-horizon objectives. However, conditioning rewards on instructions introduces a fundamental failure mode as Bellman updates couple value estimates across instruction contexts, leading to inconsistent values when instructions interrupt macro-actions. We propose \textit{Macro-Action Value Correction for Instruction Compliance} (MAVIC), which fixes Bellman backups at instruction boundaries by correcting the incoming instruction objective and restoring the continuation value under the current one. Unlike reward shaping, MAVIC modifies the bootstrapping target itself, enabling consistent value estimation under stochastic instruction switching within a unified policy. We provide theoretical analysis, an actor-critic implementation, and show that MAVIC achieves high instruction compliance while preserving base task performance in increasingly complex cooperative multi-agent environments.
\end{abstract}
% \begin{figure}[h]
%     \centering
%     \vspace{-0.8em}\
%     \includegraphics[width=0.6\textwidth]{images/introductionVC.png}
%     \vspace{-0.5em} \
%     \caption{\small Instruction following example for MAVIC in the Overcooked environment that shows that without value cancellation, the agents would not learn to follow the instruction}
%     \label{fig:value_cancel_graph}
%     \vspace{-1.5em} 
% \end{figure}

%===============================================================================
\vspace{-0.5\baselineskip}
\section{Introduction}
\vspace{-0.5\baselineskip}
Robots operating in partially observable real-world environments must often contend with dynamic natural language instructions from humans.  
These instructions arrive unpredictably, interrupt ongoing behavior, and may conflict with long-horizon objectives.
For example, a robot executing a multi-step cooking task may receive commands such as ``bring me the tomato,'' requiring immediate interruption and re-planning to complete the instructions.
In other cases, an instruction such as ``don't use the left cutting board'' alters the constraints and lead to a different optimal policy.
These challenges are amplified in cooperative multi-agent settings, where agents must maintain coordinated behavior while adapting to human inputs. 
As multi-agent systems are increasingly deployed in human-centered environments, handling such interrupting instructions is a fundamental requirement.

Various approaches have been developed to attempt to address this setting, but each with its own limitation: (i) 
Symbolic planning methods based on predefined operators (e.g., PDDL, LTL~\citep{tellex_interpreting_2021, liu_grounding_2023}) lack flexibility under uncertainty; (ii) Vision-language models (VLMs) enable instruction following~\citep{shi_hi_2025}, but are computationally expensive, usually limited to single-agent settings, and struggle with long-horizon coordination.
In contrast, multi-agent reinforcement learning (MARL) with macro-actions provides a scalable framework for long-horizon cooperation, but assumes uninterrupted execution and cannot handle dynamically arriving instructions ~\citep{amato_planning_2014,iros_macro2026}.
% We modeled the instructions as additional context to the environment state.
A natural solution is to condition rewards on these instructions by rewarding them. 
However, we show that this introduces a fundamental failure mode as Bellman updates couple value estimates across instruction contexts. When instructions interrupt macro-actions, value functions bootstrap from states governed by different objectives, causing updates to mix incompatible returns. This ``cross-contamination'' leads agents to ignore instructions or degrade base task performance, revealing a fundamental limitation of MARL under interrupting objectives.

To address this, we introduce \textit{Macro-Action Value Correction for Instruction Compliance} (MAVIC), an interruption-aware value estimation method for macro-action-based MARL. MAVIC corrects Bellman backups at instruction boundaries by subtracting the contribution of the incoming instruction objective and replacing it with the continuation value under the current task objective. This enforces conditional value consistency, ensuring each update reflects a single instruction context despite stochastic switching. Crucially, MAVIC modifies the bootstrapping target itself, yielding value estimates consistent with uninterrupted execution under each instruction while maintaining a unified policy. We provide a theoretical analysis showing that MAVIC decouples value estimation across instruction classes, making the joint problem equivalent to optimizing each instruction-conditioned objective. Based on this theory, we derive a practical actor-critic implementation that integrates with existing macro-action MARL methods and uses a lightweight frozen language encoder. %\footnote{We treat language as a stochastic conditioning signal, focusing on control-level consistency.} 
Empirically, MAVIC achieves high instruction compliance while preserving base task performance across multiple cooperative multi-agent environments, eliminating the trade-off observed in existing methods. These results highlight Bellman consistency as a key requirement for robust instruction following in MARL. Our contributions can be summarized as follows: 
\begin{itemize}[noitemsep, topsep=-3pt]
    \item Identifying cross-contamination in Bellman updates under instruction interruptions as a key failure mode in macro-action MARL.
    \item Proposing MAVIC, a value correction mechanism that enforces per-instruction consistency via modified bootstrapping.
    \item Providing a theoretical analysis showing that MAVIC decouples instruction-conditioned value functions while preserving a unified optimal policy.
    \item Empirically demonstrating that MAVIC eliminates the trade-off between instruction compliance and task performance across multiple domains.
\end{itemize}
Enabling interruption-aware learning is a critical step toward deploying multi-agent systems that can reliably integrate human input in complex real-world environments \citep{yu_asynchronous_2023,kim_openvla_2024,marl2grid_iclr2026}.

%===============================================================================
\vspace{-0.5\baselineskip}
\section{Preliminaries}
\vspace{-0.5\baselineskip}
%We consider a cooperative multi-agent setting where agents must coordinate to accomplish a shared task while operating under partial observability. This problem is often formulated as a decentralized partially observable Markov decision process (Dec-POMDP) \citep{omidshafiei_decentralized_2015}, which extends the single-agent POMDP \citep{kaelbling_planning_1998} to the multi-agent setting with a shared objective. However, many real real-world cooperative tasks, such as kitchen coordination or warehouse domains, are long horizon and complex, making learning over low level actions (e.g. joint angles) intractable. Therefore, in the following subsections we formally define our multi-agent, human instruction following problem as a macro Dec-POMDP (MacDec-POMDP).

\begin{wraptable}{r}{0.65\textwidth}
\centering
\vspace{-10pt}
\caption{MacDec-POMDP definition where the subscript $i$ denotes individual agent components.}
\vspace{-9pt}
\scalebox{0.85}{
\begin{tabular}{lll}
\multicolumn{3}{c}{}                                                                                                           \\ \hline
\multicolumn{3}{c}{\textbf{Primitive components \& functions}}                                                                                                              \\ \hline
\multicolumn{1}{l|}{\textit{Symbol}}       & \textit{Definition}                                                            & \textit{Description}                 \\ \hline
\multicolumn{1}{l|}{$S$}          & Specified by task                                                         & State space                   \\
\multicolumn{1}{l|}{$I$}          & Specified by task                                                     & Set of agents                 \\
\multicolumn{1}{l|}{$A$}          & $A = \times_{i \in I} A_i$                                             & Joint action space            \\
\multicolumn{1}{l|}{$\Omega$}     & $\Omega = \times_{i \in I} \Omega_i$                                   & Joint observation space       \\
\multicolumn{1}{l|}{$H_\Omega$}   & $\times_{i \in I} (\Omega_i \times A_i)^\mathbb{H}$                    & Joint history space           \\
\multicolumn{1}{l|}{$\gamma$}     & $\gamma \in [0,1]$                                                     & Discount factor               \\
\multicolumn{1}{l|}{$\mathbb{H}$} & $\mathbb{H} \in \mathbb{Z}$                                            & Task horizon                  \\
\multicolumn{1}{l|}{$R$}          & $R: S \times A \times S \rightarrow \mathbb{R}$         & Reward function               \\
\multicolumn{1}{l|}{$T$}          & $T: S \times A \rightarrow \Delta(S)$                   & Transition function           \\
\multicolumn{1}{l|}{$O$}          & $O: S \times A \times S \rightarrow \Delta(\Omega)$     & Observation function          \\ \hline
\multicolumn{3}{c}{\textbf{Macro components \& functions}}                                                                                                                  \\ \hline
\multicolumn{1}{l|}{$M$}          & $M = \times_{i \in I} \langle \mathcal{I}_{i,j}, \pi_{i,j}, \beta_{i,j} \rangle$ & Joint macro-action space      \\
\multicolumn{1}{l|}{$\zeta$}      & $\zeta = \times_{i \in I} \zeta_i$                                     & Joint macro-observation space \\
\multicolumn{1}{l|}{$H_\zeta$}    & $\times_{i \in I} (\zeta_i \times M_i)^\mathbb{H}$                     & Joint macro-history space     \\
\multicolumn{1}{l|}{$Z$}          & $Z: S \times A \times S \rightarrow \Delta(\zeta)$               & Macro observation function    \\ \hline

\end{tabular}}

\label{tab:mac}
\vspace{-5pt}
\end{wraptable}
We consider a cooperative MARL setting under partial observability, commonly modeled as a decentralized partially observable Markov decision process (Dec-POMDP)~\citep{omidshafiei_decentralized_2015}, which generalizes the single-agent POMDP~\citep{kaelbling_planning_1998}. In many real-world domains, tasks are long-horizon and require temporally extended actions, making learning over 1-step (primitive) actions intractable. Hence, we model the problem as a macro-action Dec-POMDP (MacDec-POMDP)~\citep{amato_planning_nodate}, where agents select temporally extended actions. A MacDec-POMDP is defined as $\mathcal{M}=\langle I, S, A, M, \Omega, \zeta, T, R, O, Z, \gamma \rangle$~\citep{amato_planning_2014} (see Table~\ref{tab:mac}).

Each agent $i$ stores its local observation $\Omega_i$ in its history $H_{\zeta,i}$ and selects macro-actions from a set $M_i$ instead of primitive actions by following a high-level policy $\Psi_i: H_{\zeta,i} \rightarrow \Delta M_i$ that maps each agent's local macro-observation history to a distribution over macro-actions. A macro-action $m_{i,j} = \langle \mathcal{I}_{i,j}, \pi_{i,j}, \beta_{i,j} \rangle$ consists of an initiation set $\mathcal{I}_{i,j} \subseteq S$, a low-level policy $\pi_{i,j}: H_\Omega \rightarrow A_i$ defined over primitive actions (i.e., common 1-step actions) that execute the actual macro-action behavior lasting for $\tau$ steps, and a termination function $\beta_{i,j}: H_\Omega \rightarrow [0,1]$. During execution, macro-actions have unknown duration a priori and unfold over multiple time steps according to $\pi_{i,j}$ until termination, at which point the agents receive a macro-observation $z_i \in \zeta_i$ and select their next macro-actions. The objective is to learn a joint high-level policy $\vec{\Psi} = \times_{i \in I} \Psi_i$ that maximizes the expected discounted return over a finite horizon, starting from an initial state $s_{(0)}$:

\begin{equation}
\max_{\vec{\Psi}} \mathbb{E}_{s_{(0)} \sim \mathcal{M}}[V^{\vec{\Psi}}(s_{(0)})]
= \mathbb{E}_{s_{(0)} \sim \mathcal{M}}\left[\sum_{t=0}^{\mathbb{H}-1}
\gamma^t R\!\left(s_{(t)}, \vec{a}_{(t)}\right) \mid s_{(0)},
\vec{\pi}, \vec{\Psi}, H_\zeta\right]
\end{equation}

% represented as $m_i = \langle I_{m_i}, \pi_{m_i},
% \beta_{m_i} \rangle$, where the initiation set $I_{m_i} \subset H_i^M$
% defines how to initiate a macro-action based on macro-observation-action
% history $H_i^M$ at the high level; $\pi_{m_i}: H_i^A \times A_i \to
% [0,1]$ is the low-level policy for achieving a macro-action, and during
% execution, the agent receives a primitive-observation $o_i \in
% \Omega_i$ based on the observation function $O_i(o_i, a_i, s) =
% P(o_i \mid a_i, s)$ at every time step; and $\beta_{m_i}: H_i^A \to
% [0,1]$ is a stochastic termination function that determines how to
% terminate a macro-action based on primitive-observation-action history
% $H_i^A$ at the low level. The objective of solving MacDec-POMDPs with
% finite horizon is to find a joint high-level policy
% $\vec{\Psi} = \times_{i \in I} \Psi_i$ that maximizes the value
% $V^{\vec{\Psi}}(s_{(0)}) = \mathbb{E}\left[\sum_{t=0}^{\mathbb{H}-1}
% \gamma^t r\!\left(s_{(t)}, \vec{a}_{(t)}\right) \mid s_{(0)}, \vec{\pi},
% \vec{\Psi}\right]$, where $\gamma \in [0,1]$ is the discount factor and
% $\mathbb{H}$ is the number of (primitive) timesteps until the problem
% terminates (the horizon).

%EXAMPLE usage of bigtimes instead of prod
% $A = \mybigtimes_{i \in I} A_i$

%===============================================================================

\vspace{-0.5\baselineskip}
\section{Macro-Action Value Cancellation for Instruction Compliance}
\vspace{-0.5\baselineskip}
% Write some stuff here like "in this section we propose our unified approach towards..." make it like 2 sentences

%\paragraph{Modeling Human Instructions in MacDec-POMDPs.}

We model human instructions in MacDec-POMDPs as elements of a discrete set of instruction classes $C$, where each class $c \in C$ corresponds to a set of semantically equivalent natural language commands (e.g., ``get me a tomato'' and ``please grab a tomato'' belong to the same class). We include a null instruction class $c_\emptyset = \{ `` \; " \}$ indicating the absence of instructions.

To incorporate instructions into the MacDec-POMDP framework, we augment the state and observation spaces with the current instruction. Given a base MacDec-POMDP $\mathcal{M}$, we define an instruction-augmented process $\mathcal{M}_C = \langle I, S \times C, A, M, \Omega \times C, \zeta \times C, T', R', O', Z', \gamma \rangle$, where the instructions modulate the reward function. In detail, each instruction class $c \in C$ induces a reward function $R_c$, and the modified reward is given by $R'((s,c), a, (s',c')) = R_c(s,a,s')$. Instruction dynamics are governed by a transition function $T'$ that jointly evolves the environment state and instruction as
$T'((s,c), a, (s',c')) = T(s,a,s') \cdot P(c' \mid c, h, a)$, where $P(c' \mid c, h, a)$ models the transition probability of instructions based on the primitive history $h$. Observations are augmented by appending the current instruction, and we force macro-action termination with instruction transitions: when a new instruction is received or an instruction terminates, the currently executing macro-action is interrupted. This coupling enables agents to respond immediately to instructions changes. Finally, we define the learning objective as optimizing expected return under each instruction class to allow agents to seamlessly and optimally solve different human instruction tasks:
\begin{equation}
\max_{\vec{\Psi}} \mathbb{E}_{s_{(0)} \sim \mathcal{M}_C}[V^{\vec{\Psi}}(s_{(0)})]
= \mathbb{E}_{s_{(0)} \sim \mathcal{M}_C}\left[\sum_{t=0}^{\mathbb{H}-1}
\gamma^t R_c\!\left(s_{(t)}, \vec{a}_{(t)}\right) \mid s_{(0)}, \vec{\pi}, \vec{\Psi}, H_\zeta\right],
\label{eq:obj}
\end{equation}
This prevents the system from biasing its behavior in anticipation of certain instructions or from ignoring some instructions that would be otherwise suboptimal under different, more common instructions.

\vspace{-0.5\baselineskip}
\subsection{Coupling Value Estimates Across Instruction Classes}\label{sec:subopt}
\begin{figure}[t]
    \centering
    \includegraphics[width=0.7\textwidth]{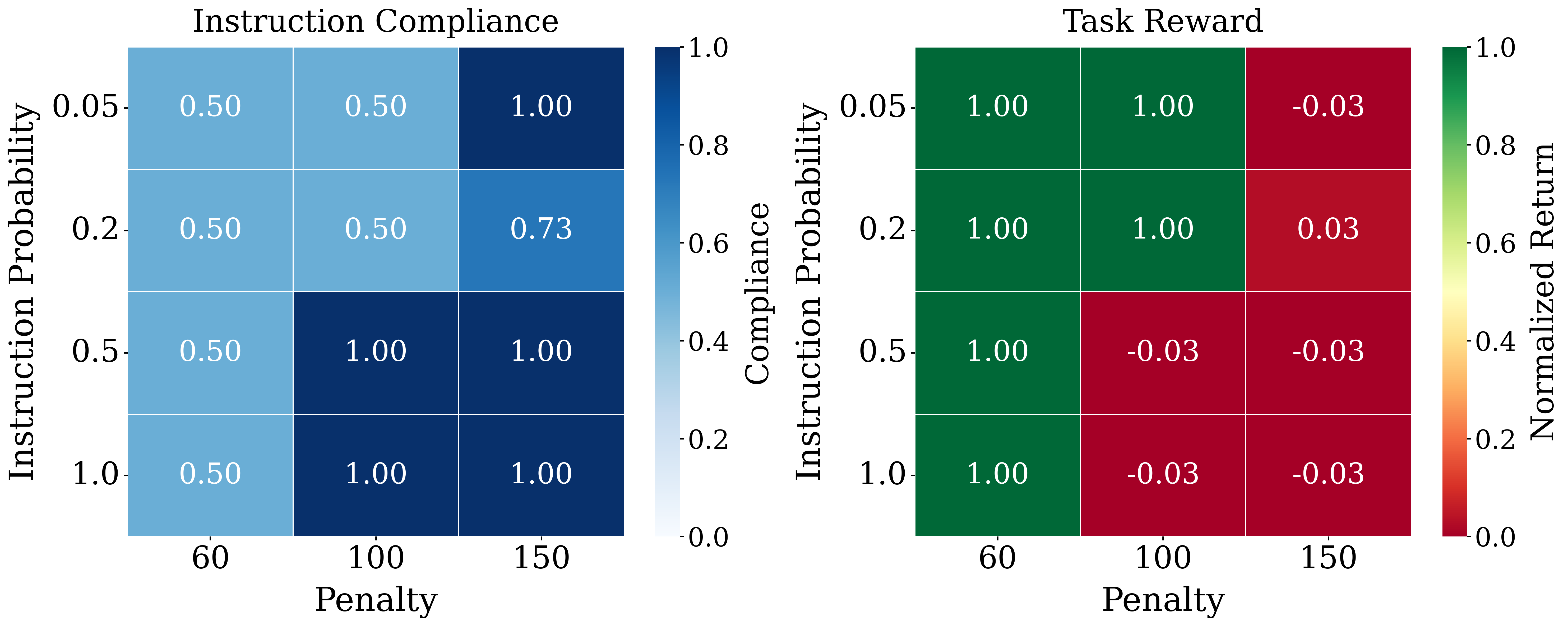}
    \caption{Demonstration of reward cross-contamination in the Box Pushing environment. With a naive reward implementation there is no penalty/probability setting which achieves both high compliance and high task reward.}
    \label{fig:heatmap}
\end{figure}
\vspace{-0.5\baselineskip}
The naive approach to follow instructions in macro-action MARL is to directly optimize the objective in Equation~\ref{eq:obj} by conditioning rewards on the current instruction. However, this naive formulation introduces an unintended coupling between instruction classes, leading to suboptimal learning and potentially degraded performance.\citep{marl_credit_aamas}. For example, consider a simplified setting with two instruction classes: a null instruction $c_\emptyset$ and a single active instruction $c$. Even when evaluating the value of a policy in a non-instruction state $(s, c_\emptyset)$, the expected return depends on the probability of transitioning into an instruction state and the policy’s performance under that instruction. Concretely, denoting by $\beta$ the probability of receiving an instruction, the value function takes the form:
\begin{equation}
V^{\vec{\Psi}}((s, c_\emptyset)) = \mathbb{E}\left[ R_{c_\emptyset} + \gamma \left( \beta \, V^{\vec{\Psi}}((s', c)) + (1 - \beta) \, V^{\vec{\Psi}}((s', c_\emptyset)) \right) \right],
\end{equation}
This highlights that the value in non-instruction states is recursively coupled with its instructions value, and optimizing performance for one instruction class can adversely affect performance. Agents may learn to ignore instructions that are suboptimal with respect to the base task, or alter their base behavior in anticipation of future instructions, leading to degraded overall performance. %Specifically, in standard MARL the Bellman update bootstraps from the value under the new instruction $V((s', c))$. As a result, the return used to update the original macro-action includes contributions from a different objective, causing value estimates for $c_\emptyset$ to depend on outcomes under $c$.
We refer to this phenomenon as \emph{cross-contamination} of value estimates across instruction classes. Figure~\ref{fig:heatmap} demonstrates cross-contamination empirically in the Box Pushing environment (described in Section \ref{sec:result}) with a single instruction class ``do not push any boxes'' resulting in a fixed penalty if the instruction is ignored. Figure~\ref{fig:timeline} additional illustration of this failure mode (top red row), highlighting how cross-contamination propagates through the Bellman objective used in standard MARL methods.

 % We see, under a naive reward implementation, that there are no combinations of $\beta_\emptyset$ and instruction violation penalty which achieve both high levels of base task performance and instruction compliance. 

% Figure~\ref{fig:timeline} illustrates this failure mode (top red row). Consider a trajectory in which an agent is executing a macro-action under instruction $c$, and a new instruction $c'$ arrives mid-execution, forcing early termination. In standard MARL, the Bellman update bootstraps from the value under the new instruction $V((h', c'))$. As a result, the return used to update the original macro-action includes contributions from a different objective, causing value estimates for $c$ to depend on outcomes under $c'$. This leads to cross-contamination across instruction contexts. 

%MAVIC corrects this at the update level (bottom green row). When an instruction change occurs, the contribution of the incoming instruction is replaced with the continuation value under the current instruction, $V((h', c))$. This ensures that the update reflects only the return that would have been obtained had the instruction remained unchanged. Consequently, each value function is trained on consistent targets within its own instruction context, eliminating cross-contamination with a unified policy.
\vspace{-0.5\baselineskip}
\subsection{Value Decoupling for Instruction-Conditioned MARL}\label{sec:rew}
\begin{figure}[b]
    \centering
    \vspace{-10pt}\includegraphics[width=0.9\textwidth]{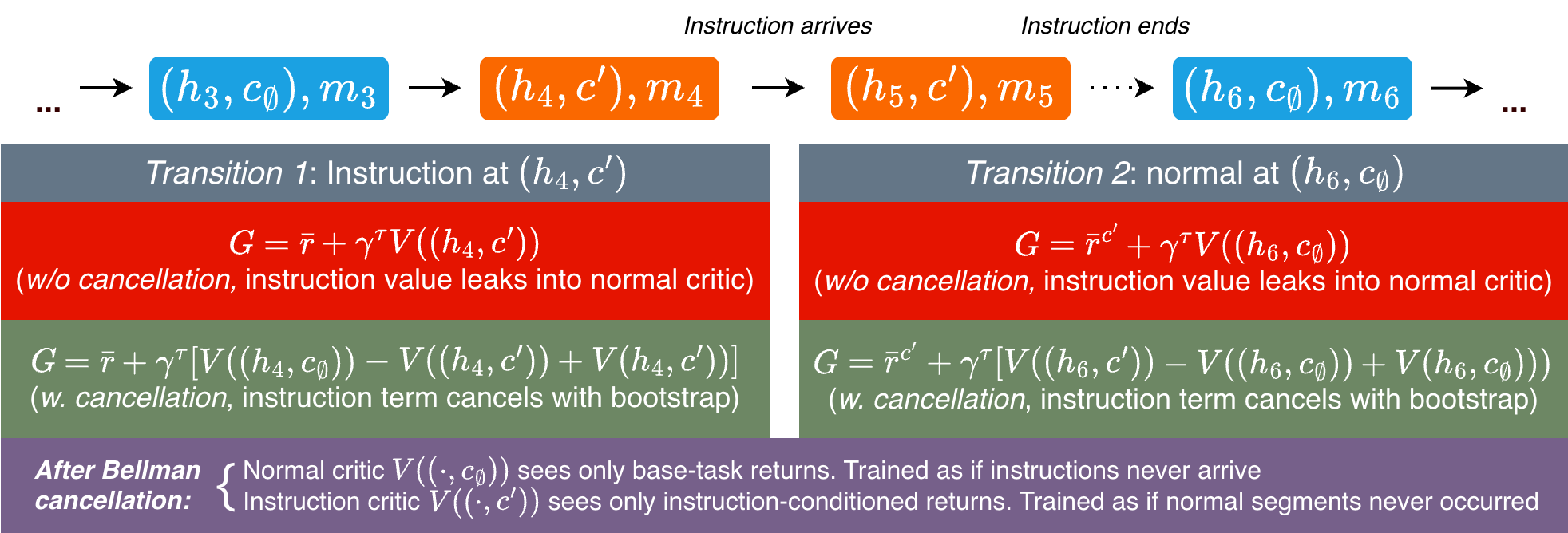}
    \caption{Illustration of value cross-contamination and MAVIC correction. Top (red): standard Bellman updates bootstrap from the value under a new instruction after interruption, causing value estimates to mix across instruction contexts. Bottom (green): MAVIC corrects this contribution and restores the continuation value under the original instruction, ensuring consistent value updates.}
    \label{fig:timeline}
    \vspace{-5pt}
\end{figure}

Our key idea is to correct Bellman updates at instruction transitions so that value estimates remain consistent within each instruction class (Figure~\ref{fig:timeline}, bottom green). When an instruction change occurs, the contribution of the incoming instruction is replaced with the continuation value under the current instruction, $V((h', c))$. This ensures that the update reflects only the return that would have been obtained had the instruction remained unchanged. Consequently, each value function is trained on consistent targets within its own instruction context, eliminating cross-contamination with a unified policy. In contrast, the naive Bellman update bootstraps as $V((h', c'))$, causing value estimates to depend on outcomes from different instruction contexts. MAVIC removes this dependence, decoupling value estimation across instructions while maintaining a shared policy.

Without correction, Bellman updates mix returns across instructions, violating per-instruction value consistency; MAVIC restores this property through value correction. The following theoretical results consider ($\mathcal{M}_C$) under the value-corrected reward.
%\textbf{Why this correction is necessary.}
%Without correction, Bellman updates for a given instruction recursively depend on value estimates under other instructions, mixing returns across objectives. Consequently, the value of a policy is not well-defined for any fixed instruction, as it depends on trajectories governed by different objectives. In other words, the Bellman operator fails to preserve per-instruction semantics. MAVIC restores this property by enforcing consistency with uninterrupted execution under the current instruction.

\textbf{Lemma 1.} \textit{The value of a policy $\Psi$ in $\mathcal{M}_C$ given an instruction class $c \in C$ is equivalent to its value in the instruction-specific problem $\mathcal{M}_c$ for all $s \in S$.}

To enforce this property, we define a dynamic reward $R'$ (Table~\ref{tab:dynamic_rew}) that augments $R_c$ at instruction transitions. When $c = c'$, the reward is unchanged. When $c \neq c'$, the correction subtracts the bootstrap from the incoming instruction and replaces it with the continuation value under the current instruction. This ensures that future instruction transitions do not influence the value of the current instruction, eliminating cross-contamination.

\begin{table}[h]%{0.65\textwidth}
\centering
\vspace{-10pt}
\caption{Dynamic reward function $R'((s,c), \vec{a}, (s',c'))$}
\label{tab:dynamic_rew}
\begin{tabular}{cl}
\toprule
\multicolumn{1}{c|}{\textbf{Transition}} & \textbf{Reward} \\ \midrule
\multicolumn{1}{c|}{$c = c'$} & $R_c(s,\vec{a},s')$ \\
\multicolumn{1}{c|}{$c \neq c'$} & $R_c(s, \vec{a}, s') - \gamma V^{\vec{\Psi}}((s', c')) + \gamma V^{\vec{\Psi}}((s', c))$ \\ \bottomrule
\end{tabular}
\vspace{-5pt}
\end{table}

\paragraph{Relation to reward shaping.}
While the correction may resemble potential-based shaping~\citep{ng1999policy}, it is fundamentally different. Potential-based methods add a fixed term $\gamma \Phi(s') - \Phi(s)$ and preserve the Bellman operator. In contrast, MAVIC introduces a \emph{policy-dependent} correction that directly modifies the bootstrapping target. It therefore cannot be expressed as a fixed potential function and instead enforces consistency at the level of Bellman updates under stochastic instruction switching. This distinction is critical under interruptions: when an instruction arrives mid-execution, the naive update bootstraps from a different instruction context, introducing bias and training on truncated behaviors. MAVIC instead bootstraps from the intended continuation, preserving consistency with the original macro-action.

\textbf{Theorem 1.} \textit{If a policy $\Psi^*$ is optimal in $\mathcal{M}_C$, then it is also optimal in each instruction-specific problem $\mathcal{M}_c$ for all $c \in C, s \in S$.}

This decoupling arises solely from the modified Bellman updates, without restricting transitions or policies. More broadly, the result applies to settings with stochastic objective switching and temporally extended actions beyond language-conditioned MARL, which is an exciting direction for future investigation.

%In the naive formulation, value estimates for one instruction are recursively influenced by others, preventing consistent optimization within any single instruction context. In contrast, under the modified reward, each instruction-conditioned value function evolves independently while sharing a common policy. This decoupling explains why MAVIC achieves both high instruction compliance and strong base-task performance.

\subsection{MAVIC: Learning with Value Correction}
\label{sec:mavic}
\vspace{-0.5\baselineskip}

The value-corrected reward formulation (Table~\ref{tab:dynamic_rew}) provides a principled solution, but requires approximating instruction-conditioned value functions under partial observability and natural language inputs. We address this with a history-conditioned actor-critic architecture and a training procedure that applies value correction at instruction transitions, which can be implemented on top of existing macro-action actor-critic methods such as the asynchronous framework of \citet{xiao_asynchronous_2022}.

\begin{wrapfigure}{r}{0.6\textwidth}
    \centering
    \includegraphics[width=\linewidth]{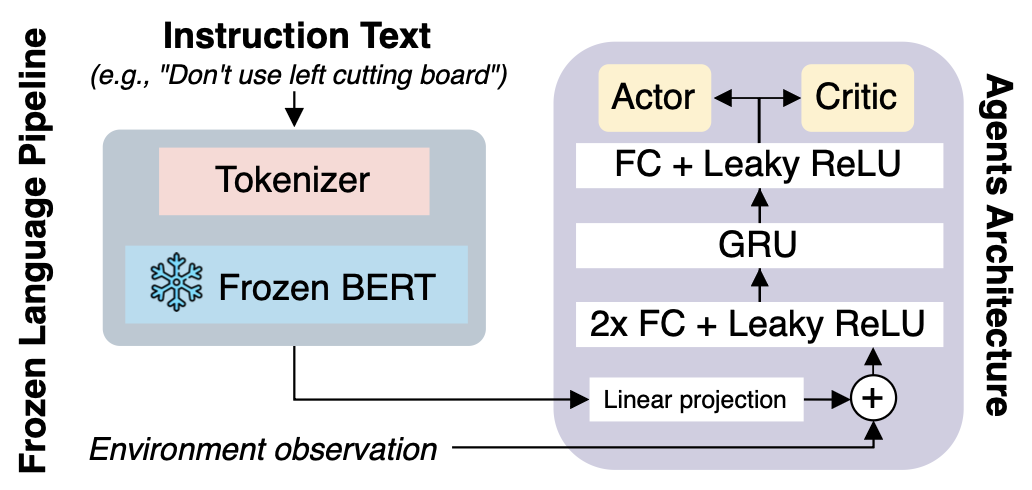}
    \caption{The actor and critic are modified to take the instructions given as natural language as input. This is done through embedding each instruction with a frozen BERT model \citep{devlin_bert_2019} which is then concatenated with the observation later on.}
    \label{fig:actor_critic}
\vspace{-7pt}
\end{wrapfigure}
Figure~\ref{fig:actor_critic} illustrates the MAVIC architecture. Each agent maintains an actor network $\Psi_{\theta_i}$ (where $\theta_i$ parametrizes agent's $i$ policy) that selects macro-actions conditioned on its macro-observation history and the current instruction.

Natural language instructions are encoded using a frozen language encoder (BERT \citep{devlin_bert_2019}), and the resulting embedding is projected down and concatenated with the agent’s history before being processed by the network. To support value correction, each agent maintains instruction-conditioned critics $V_{\phi_i}((h, c))$ (parametrized by $\phi_i$) that estimate the value of a history $h$ under instruction class $c$. In practice, these critics share parameters but are conditioned on the instruction embedding, enabling generalization across different phrasings of the same instruction. This design replaces state-based value functions $V((s, c))$ of Theorem 1 with history-conditioned approximations $V((h, c))$, which are standard in partially observable settings and serve as sufficient statistics for decision making~\citep{baisero_unbiased_2022}.

\paragraph{Algorithm.} The MAVIC training procedure is summarized in Algorithm~\ref{alg:mavic}. During data collection (lines 3-9), agents execute macro-actions until either a macro-action terminates or a new instruction arrives (line 5). When a macro-action terminates naturally, only the corresponding agent selects a new macro-action. Instruction transitions are sampled according to $\beta_c$ (line 6), and when a new instruction is received, all active macro-actions are terminated and new macro-actions are selected conditioned on the updated instruction (line 7). Transitions at time $t$ are stored in a replay buffer as tuples $(h_t, c_t, h_{t+1}, c_{t+1}, \bar{r}_t, \vec{m}_t)$, capturing both environment evolution and instruction changes.

\begin{algorithm}[t]
\caption{MAVIC Training Template}
\label{alg:mavic}
\begin{algorithmic}[1]
\Require $\mathcal{M}$, instruction classes $C$, frozen language encoder $\mathcal{E}$
\State Initialize actors $\Psi_{\theta_i}$, critics $V_{\phi_i}$, replay buffer $\mathcal{D}$
\For{episode $= 1, 2, \ldots$}
    \State Sample $s_0 \sim \mathcal{M}_C$, set $c_0 = c_\emptyset$, sample $\vec{m} \sim \vec{\Psi}(\cdot \mid h_0, \mathcal{E}(`` \; "))$
    \While{episode not terminated}
        \State Execute primitives from $\vec{m}$ until any macro-action terminates; append to $\mathcal{D}$
        \State With prob.\ $\beta_c$: terminate all $\vec{m}$, sample $c' \sim \beta_c$, $t \sim c'$
        \State Sample $\vec{m} \sim \vec{\Psi}(\cdot \mid h, \mathcal{E}(t))$
    \EndWhile
\EndFor
\For{training epoch $= 1, 2, \ldots$}
    \State Sample batch $\{(h_k, c_k, h_{k+1}, c_{k+1}, \bar{r}_k, \vec{m}_k)\} \sim \mathcal{D}$
    \If{$c_k \neq c_{k+1}$}
        \State $\bar{r}_k \gets \bar{r}_k + \gamma^{\tau_m} [V_{\phi}((h_{k+1}, c_k)) - V_{\phi}((h_{k+1}, c_{k+1}))]$
    \EndIf
    \State Compute $G$ via Eq.~\ref{eq:return}; update $\Psi_{\theta_i}$, $V_{\phi_i}$ via provided actor-critic gradients
\EndFor
\end{algorithmic}
\end{algorithm}

During training (lines 10-15), the key step is the application of value correction. For each sampled transition, if an instruction change occurs ($c_t \neq c_{t+1}$), we apply the correction (line 12) where $\tau_m$ is the macro-timestep:
\begin{equation}
\bar{r}_t \leftarrow \bar{r}_t + \gamma^{\tau_m} \left[ V_{\phi}((h_{t+1}, c_t)) - V_{\phi}((h_{t+1}, c_{t+1})) \right],
\end{equation}
which removes the contribution of the incoming instruction and replaces it with the continuation value under the current instruction. Using this corrected reward, we compute the return $G$ over macro-action segments (line 14):
\begin{equation}
\label{eq:return}
G = 
\begin{cases}
    \bar{r}^{c} + \gamma^{\tau_m} V_{c}
        & \textit{if}~c = c', \\[4pt]
    \bar{r}^{c} + \gamma^{\tau_m} (V_{c} - V_{c'}) + \gamma^{\tau_m}V_c'
        & \textit{if}~c \neq c'.
\end{cases}
\end{equation}
% \begin{equation}
% \label{eq:return}
% G = 
% \begin{cases}
%     \bar{r} + \gamma^{\tau_m} V_{c_\emptyset}
%         & \textit{if}~c = c' = c_\emptyset, \\[4pt]
%     \bar{r} + \gamma^{\tau_m} (V_{c_\emptyset} - V_{c'})
%         & \textit{if}~c = c_\emptyset,\, c' \neq c_\emptyset,
% \end{cases}
% \quad
% \begin{aligned}
%     &\bar{r}^{c'} + \gamma^{\tau_m} V_{c'}
%         \quad \textit{if}~c = c' \neq c_\emptyset, \\[4pt]
%     &\bar{r}^{c} + \gamma^{\tau_m} (V_{c} - V_{c_\emptyset})
%         \quad \textit{if}~c \neq c_\emptyset,\, c' = c_\emptyset.
% \end{aligned}
% \end{equation}

This value correction mechanism provides a practical realization of Theorem 1. By removing the contribution of future instruction states from each update, the corrected return ensures that the value function is trained as if the instruction class $c$ remained fixed. Consequently, each instruction-conditioned value function is optimized independently, allowing the learned policy to recover per-instruction optimal behavior while remaining unified across all instruction classes.

To implement MAVIC within an actor-critic framework, we derive a macro-action policy gradient that is consistent with the value-corrected objective, ensuring that policy updates respect the decoupled structure established in Theorem 1. The resulting gradient for agent $i$ is given by:
\begin{equation}
\nabla_{\theta_i} J(\theta_i) = \mathbb{E}_{\vec{\Psi}_{\vec{\theta}}}\left[\nabla_{\theta_i} \log \Psi_{\theta_i}(m_i \mid h_i, c) \left(\bar{r}^{\,c}_i + \gamma^{\tau_{m_i}} V^{\vec{\Psi}}((h'_i, c')) - V^{\vec{\Psi}}((h_i, c))\right)\right].
\label{eq:advantage}
\end{equation}
The full derivation is provided in Appendix~\ref{app:theorems}. This gradient mirrors the standard macro-action policy gradient, with $\bar r_i^c$ denoting the corrected macro-reward from Equation~\ref{eq:return}; therefore the complete target is conditioned only on the current instruction context. Then,
each critic is trained by minimizing the temporal-difference error $\mathbb{E} \left[ \left( G - V_{\phi}((h, c)) \right)^2 \right],$
and the actor is updated using the policy gradient method with the corrected returns (line 15). As illustrated in Figure~\ref{fig:timeline}, this procedure ensures that Bellman updates no longer bootstrap across instruction boundaries, preventing value leakage and enabling stable learning under temporally interleaved instructions.

\paragraph{Convergence and approximation scope}
\label{app:convergence}
MAVIC inherits the convergence assumptions and limitations of the underlying asynchronous macro-action actor-critic framework~\citep{xiao_asynchronous_2022}. The correction itself introduces no additional cross-instruction bootstrap error: because the same critic estimate appears in the cancellation and in the ordinary bootstrap, the incoming-context term cancels exactly as shown in Equation~\ref{eq:return}. Hence, whenever the underlying actor-critic converges for an instruction-specific objective, MAVIC reconstructs that same learning target at instruction boundaries; Theorem~1 then establishes the corresponding optimality-preservation property across instruction classes. With nonlinear function approximation, we do not claim a stronger global convergence guarantee than that of the underlying actor-critic method.

\paragraph{Limitations.}
MAVIC relies on accurate estimation of instruction-conditioned value functions, which can be challenging under partial observability and sparse instructions, potentially increasing variance in the correction terms. It also assumes that instruction embeddings generalize across semantically equivalent commands, which may degrade under diverse or ambiguous language. Nonetheless, MAVIC focuses on control-level consistency, not language grounding. Finally, while MAVIC mitigates interference between instruction classes, it does not explicitly address exploration over rare or imbalanced instructions. These limitations primarily reflect opportunities for improving value estimation, representation learning, and exploration in instruction-conditioned MARL.

%===============================================================================

\vspace{-0.5\baselineskip}
\section{Experiments}
\label{sec:result}
\vspace{-0.5\baselineskip}
\paragraph{Environments.}

We evaluate the performance of MAVIC with the box pushing (BP), warehouse tool delivery (WH) \citep{xiao_asynchronous_2022}, and the overcooked (OC) \citep{wu_coordinated_2021} domains, which are standard benchmarks in the MARL macro-action literature. Macro-actions are predefined in both domains and the primitive actions are part of the set of macro-actions. In the following, we briefly describe the environments, referring to Appendix \ref{app:implementation} and the original works for more details:

\textbf{BP}: Two robot agents must coordinate to reach and push a central box to a target location. Instructions include: ``Go to small boxes'' and ``Don't push the box.''\newline
\textbf{WH}: Three heterogeneous robot agents (two mobile, one arm) assist three human workers by delivering tools whose required order may change based on instructions. Humans move between work and break stations. Instructions include: ``Get me tool 0-3.'' \newline
\textbf{OC}: Three agents (and a human) collaborate in a grid $7\times7$ to prepare dishes that require cutting, blending, baking and delivery. Agents receive rewards for correct actions and penalties for delays or mistakes. Instructions include constraints (``Don't use the left cutting board''), suboptimal requests (``Get me the lettuce''), and direct commands (``Move left''). Other instructions will be noted in Appendix~\ref{app:environment_details}

\begin{figure}[b]
\centering
\begin{subfigure}[b]{0.31\linewidth}
  \centering
  \includegraphics[height=3cm,keepaspectratio]{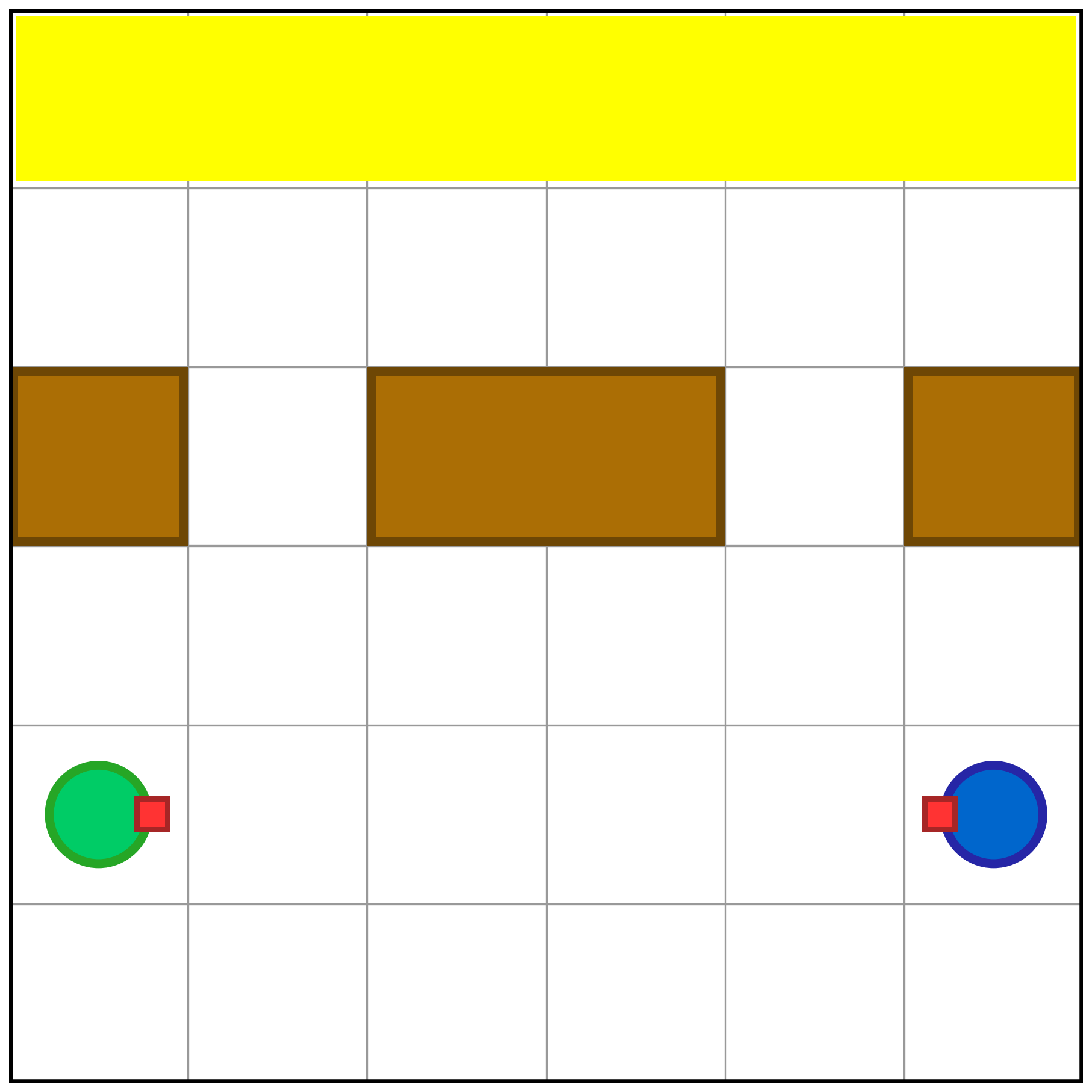}
  \caption{BP}
  \label{fig:env-bp}
\end{subfigure}
\begin{subfigure}[b]{0.31\linewidth}
  \centering
  \includegraphics[height=3cm,keepaspectratio]{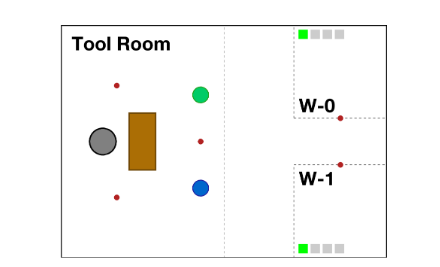}
  \caption{WH}
  \label{fig:env-wtd}
\end{subfigure}
\hspace{\fill}
\begin{subfigure}[b]{0.31\linewidth}
  \centering
  \includegraphics[height=3cm,keepaspectratio]{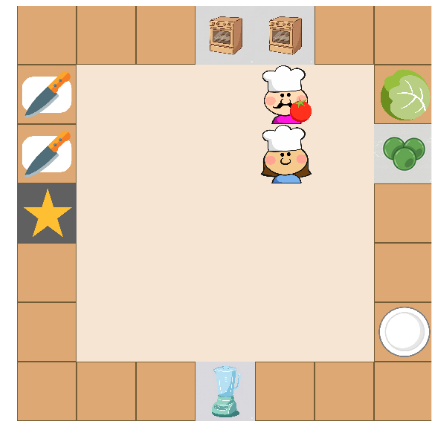}
  \caption{OC}
  \label{fig:env-oc}
\end{subfigure}
\vspace{-3pt}
\caption{Overview of the tasks. BP is Boxpushing, WH is Warehouse, and OC is Overcooked.}
\label{fig:environments}
\vspace{-10pt}
\end{figure}

% \begin{itemize}[noitemsep, topsep=-2pt]
%     \item \textit{BP}: Two robot agents must coordinate to reach and push a central box to a target location. Instructions include: ``Go to small boxes'' and ``Don't push the box.''
%     \item \textit{WTD}: Three heterogeneous robot agents (two mobile, one arm) assist three human workers by delivering tools whose required order may change based on instructions. Humans move between work and break stations. Instructions include: ``Get me tool 0-3.''
%     \item \textit{OC}: Three agents (and a human) collaborate in a grid $7\times7$ to prepare dishes that require cutting, blending, baking and delivery. Agents receive rewards for correct actions and penalties for delays or mistakes. Instructions include constraints (``Don't use the left cutting board''), suboptimal requests (``Get me the lettuce''), and direct commands (``Move left'').
% \end{itemize}

\textbf{Experiment setup and baselines.} 
We evaluate MAVIC on top of the macro-action actor-critic methods Mac-IAC and Mac-CAC, implemented using the original codebase of \citep{yu_asynchronous_2023}, with hyperparameters selected via grid search (Appendix~\ref{app:hyperparameters}). We compare MAVIC implementations against two instruction-aware baselines: (i) a naive approach directly using instruction task rewards with no value correction, similar to the top row of Figure~\ref{fig:timeline}; (ii) a ``switch'' training paradigm that gathers training data from each instruction class independently, side stepping the cross-contamination issue. Both baselines highlight different challenges in learning to optimize our task-aware objective. The naive baseline struggles with reward cross-contamination, as we discussed in Section~\ref{sec:subopt}, and our switch training baseline illuminates the difficulty of exploring unique instruction contexts and reducing training instability even when cross contamination is eliminated. Lastly, all methods are compared against vanilla Mac-IAC, trained and evaluated without instructions,  to act as an upper bound on base task performance. 
% (i) the underlying macro-action method without instruction interruptions or value correction (``vanilla'') and (ii) an interruption-aware variant that incorporates instructions but omits value correction (``baseline''). This setup isolates the causal impact of value correction across architectures with different critic structures and policy gradient formulations. We further show that failures under out-of-distribution instructions stem from representation limitations rather than value estimation.

\textbf{Metrics} We evaluate all methods across two main metrics: base task performance and instruction compliance. Base task performance is measured without instructions being given in the environment, allowing for a clean and accurate measure of how well each method learns to solve the original environment task. Compliance is measured within a live instruction environment, meaning instructions are given online at random intervals during execution. We compute compliance as the ratio of instructions followed divided by the total number of instructions given. For details on how instruction compliance is measured for each environment and instruction pair, see Appendix~\ref{app:environment_details}.
Unless otherwise noted, results correspond to the average return over a 10-episode window, aggregated across 5 independent training seeds per method with shaded regions indicating $95\%$ bootstrapped confidence intervals. 
%Experiments were run on a research cluster equipped with Xeon E5-2650 CPU nodes with 256 of RAM. 

\vspace{-0.5\baselineskip}
\subsection{Empirical Evaluation}
\vspace{-0.5\baselineskip}
\begin{table}[t]
\caption{Comparison of Base and Compliance metrics across different algorithms and environments. Best performing values for each metric are highlighted in bold.}
\centering
\resizebox{\textwidth}{!}{
\begin{tabular}{lcccccccccccc}
\toprule
\multicolumn{13}{c}{\textbf{Algorithm performance across environments}}                                                                                                                                                                                    \\ \midrule
\multicolumn{1}{l|}{\text{Environment}} & \multicolumn{4}{c|}{\textbf{OC}}                                           & \multicolumn{4}{c|}{\textbf{BP}}                                           & \multicolumn{4}{c}{\textbf{WH}}                      \\ \midrule
\multicolumn{1}{l|}{Metric}               & Base            & $\sigma$ & Compl.        & \multicolumn{1}{c|}{$\sigma$} & Base            & $\sigma$ & Compl.        & \multicolumn{1}{c|}{$\sigma$} & Base            & $\sigma$ & Compl.        & $\sigma$ \\ \midrule
\multicolumn{1}{l|}{\textbf{MAVIC}}       & 122.28 & 0.56     & 99\% & \multicolumn{1}{c|}{1.3\%}    & 288.32 & 4.1      & 99\% & \multicolumn{1}{c|}{0.14\%}     & 438.90          & 15.238   & 93\% & 2.0\%    \\
\multicolumn{1}{l|}{\textbf{Naive}}       & 3.62            & 9.98     & 93\%          & \multicolumn{1}{c|}{3.3\%}    & 270.60          & 5.68     & 5\%           & \multicolumn{1}{c|}{5.0\%}    & 359.12          & 63.62    & 92\%          & 1.3\%    \\
\multicolumn{1}{l|}{\textbf{Switch}}      & 104.00          & 49.46    & 99\% & \multicolumn{1}{c|}{0.1\%}    & 267.85          & 3.90     & 6\%           & \multicolumn{1}{c|}{7.9\%}    & 454.49 & 30.11    & 86\%          & 21\%     \\ \midrule
\multicolumn{1}{l|}{\textbf{Vanilla Mac-IAC}}     & 152.31          & 22.00    & N/A           & \multicolumn{1}{c|}{N/A}      & 290.42          & 0.01     & N/A           & \multicolumn{1}{c|}{N/A}      & 473.65          & 13.35    & N/A           & N/A      \\ \bottomrule
\end{tabular}
}
\label{tab:performance_comparison_wide}
\end{table}

% We first empirically demonstrate the value coupling limitation described in Section~\ref{sec:subopt} in the box pushing domain. Figure~\ref{fig:heatmap} shows the results for increasing instruction activation probability and penalty magnitude for the baseline (i.e., without value correction), where the instruction remains active for the rest of the episode once triggered. At low penalty (e.g., 60), instruction compliance is low while task reward remains high, indicating that the agent ignores the instruction. As the penalty increases, compliance rises to near 1.0, but task reward collapses to near zero, as the policy shifts to avoid penalties at the expense of task completion. \textit{Notably, no configuration in the grid achieves both high compliance and high task reward, illustrating the fundamental trade-off induced by value coupling.}

In Table~\ref{tab:performance_comparison_wide} we compare the performance of MAVIC against the naive and switch baselines across our three evaluation environments. Our results show that MAVIC is the only method consistently capable of achieving both high levels of base task performance and instruction compliance. Specifically, our results for the naive baseline gives further proof of the cross contamination problem we discussed in Section~\ref{sec:subopt} as the method either demonstrates a significant drop in base task performance, in the case of OC and WH, or a complete lack of task compliance, as in BP. 

Additionally, our switch baseline also struggles in this setting despite being trained independently between instruction and base task instances, showing either lower levels of compliance than MAVIC,  in the case of BP and WH, or a drop in base task performance, as in OC. This is a likely result of unstable training induced by per-episode task switching, which is further evidenced by the high standard deviation values for the method seen in OC and WH. Furthermore, since the agents remain within a single task context for each episode under the switch training paradigm, they are less likely to explore a diverse set of initial instruction contexts. For instance, if the agents start with the instruction ``don't move'', they will never explore other states outside the initial state set once they learn the instruction.

In contrast, MAVIC trains directly within the instruction interruption environment, meaning it can learn how to dynamically switch between tasks while simultaneously prevent cross-task reward contamination. As a result, MAVIC is a more principled approach for training instruction-aware agents by enabling in-context training across all instructions. Additional interruption-recovery, switch-frequency, critic-noise, transition-noise, and reward-shaping analyses are reported in Appendix~\ref{app:robustness}.

\textbf{Is value correction necessary?} To isolate the contribution of value correction, we compare MAVIC against a zero-bootstrap variant that removes the incoming instruction's bootstrap at instruction boundaries without restoring the continuation value under the original instruction. Table \ref{tab:zero_bootstrap} shows that both methods achieve comparable base-task returns in WH. However, their performance under online interruptions differs substantially: zero bootstrap achieves only 4.4\% episode completion, compared to 96.9\% for MAVIC, and its post-switch return drops from 264.4 to -17. These results demonstrate that eliminating cross-instruction value contamination alone is insufficient. Restoring the continuation value is crucial for preserving coordinated behavior across instruction boundaries.

\begin{table}[b]
\vspace{-12pt}
\centering
\small
\caption{Zero-bootstrap ablation in Warehouse.}
\label{tab:zero_bootstrap}
\begin{tabular}{lccccc}
\toprule
Method & Base return & Recovery & Completion & Post-switch return & Waiting \\
\midrule
Zero bootstrap & $420.1\pm1.4$ & $87.8\%$ & $4.4\pm6.3\%$ & $-17\pm23$ & $230.3\pm62.6$ \\
MAVIC & $423.4\pm28.7$ & $100\%$ & $96.9\pm4.4\%$ & $264.4\pm46.3$ & $130\pm24.6$ \\
\bottomrule
\end{tabular}
\end{table}

\vspace{-0.5\baselineskip}
\subsection{Environment Specific Case Studies}
\vspace{-0.5\baselineskip}

\textbf{Overcooked} OC is notable for having high levels of compliance across all methods while seeing the most significant drops in base task performance in the naive and switch baselines. Specifically, under restrictive instructions like ``don't use the left cutting board'' or ``don't get the tomato'' the agents receive a large negative reward (e.g., $-50$) for accessing the restricted resource. This greatly biases returns received by agents trained with naive rewards, resulting in them avoiding the tomato and left cutting board even when the instructions are inactive. Since these resources are necessary for base task completion, the agents never solve the base task. 

We see a similar drop in base task performance for the switch baseline in OC, though not as severe, along with a high level of variance across training seeds. This variance is caused by the switch baseline failing to learn the base task across a number of training seeds. In particular, we noticed a similar failure mode as observed in the naive baseline, the agents learn to always follow restrictive instructions even when they are not given. It is possible that this issue could be solved by adjusting how often the method switches between environment task types, however this meta optimization is completely unnecessary with MAVIC which shows much higher levels of stability across all environments and instructions.

\textbf{Boxpushing}
The BP environment has proven to be difficult for all methods to comply with the instruction as the environment is more difficult to solve compared to WH and OC. On BP, complying with "push the small boxes" requires both agents to abandon the dominant +300 big-box reward which is different from OC and WH where compliance is usually a local sub-task swap. Naive and Switch cannot resolve the reward ambiguity and collapse to $5\%$ compliance focusing on task performance, treating the instruction input as noise. MAVIC, however, is able to separate task and instruction cleaner reaching $99\%$ compliance while preserving Vanilla's benign return (288 vs 290).

\begin{figure}[b]
    \centering
    \vspace{-12pt}
\includegraphics[width=1.0\textwidth]{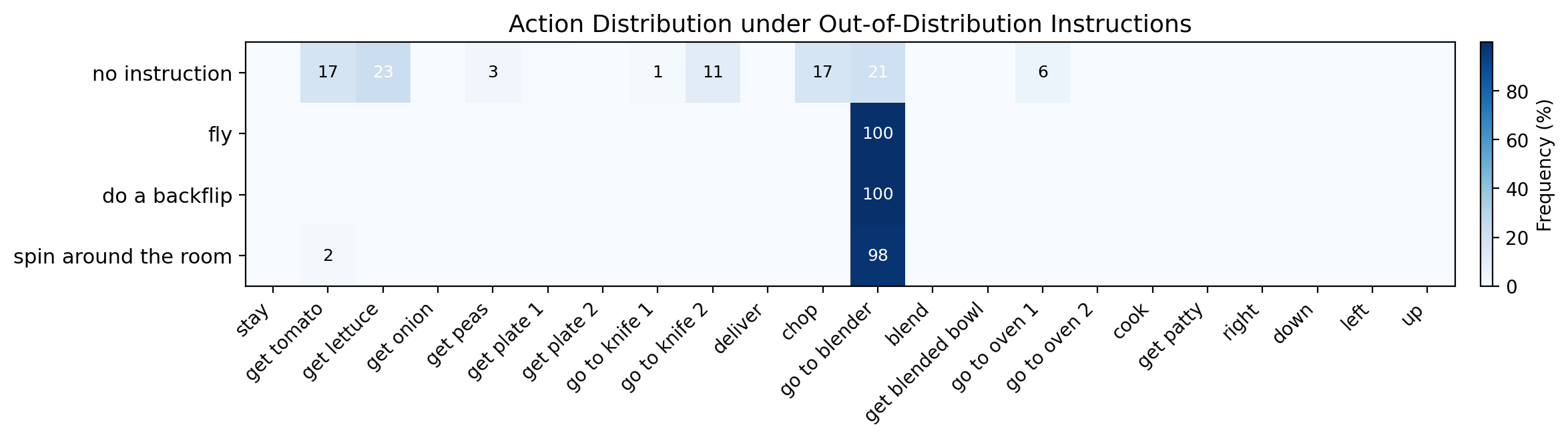}
    \vspace{-10pt}
    \caption{Macro-action frequencies under no instruction and infeasible out-of-distribution ones.}
    \label{fig:ablation_action_dist}
    \vspace{-5pt}
\end{figure}

\textbf{Warehouse tool delivery} Figure~\ref{tab:performance_comparison_wide} shows the results in the WH domain and confirm our previous takeaways.  When instructions only reorder existing rewards (e.g., by changing the tool delivery order in WH) without introducing additional shaping terms, MAVIC and the other baselines perform well. In this setting, instruction and base-task objectives remain aligned, so value coupling does not significantly impact learning. However, changing the tool delivery order in the middle of the episode makes the task harder, and MAVIC successfully delivers the new tool correctly with low variance while the other methods struggle. Despite similar rewards, the baselines have a tradeoff by having higher variance and lower compliance. These results highlight that value correction is most critical when instruction-conditioned rewards alter the underlying value structure, indicating that \textit{MAVIC is necessary in tasks where instruction signals induce conflicting value estimates.}

\textbf{Out-of-Distribution Instructions.}
Furthermore, we investigate MAVIC's performance on out of distribution instructions not observed during training. We first evaluated MAVIC on instructions semantically similar to those seen during training, but still out of distribution. We show our results in Table~\ref{tab:ood_compliance} where we observe MAVIC's ability to generalize across novel instructions. Next, we probe the behavior of MAVIC under instructions that are far out of distribution, representing irrelevant or infeasible requests (e.g., ``fly,'' ``spin around the room''). In Figure~\ref{fig:ablation_action_dist} we observe that every infeasible instruction results in a single, shared macro-action being chosen with high probability. This uniformity of behavior is reflective of the relatively limited language coverage the agents were trained on, as all instructions are mapped to identical behaviors. This emphasizes the need for calibrated language representations and a diversity of language instructions to ensure better generalization. 

% First, we test MAVIC on instructions that are physically infeasible or semantically unrelated to the training set (e.g., ``fly,'' ``spin around the room''). Figure~\ref{fig:ablation_action_dist} shows the resulting macro-action distributions, comparing against no instructions as a baseline. Notably, all infeasible instructions collapse to a single macro-action which reflects high-confidence misinterpretation rather than uncertainty, further emphasizing the need for calibrated language representations. This behavior highlights a limitation of the language representation rather than the value estimation mechanism. Because the language encoder is frozen, semantically unrelated instructions are mapped to arbitrary regions of the embedding space. 

\begin{table}[h]
\centering
\caption{Instruction Compliance for Out-of-Distribution Commands}
\label{tab:ood_compliance}
\begin{tabular}{llcc}
\toprule
OOD instruction & Intent & Compliance & $\sigma$ \\
\midrule
``the delivery is mine'' & deliver & \textbf{99.9\%} & 0.4\% \\
``i've got the right cutting board'' & knife-1  & \textbf{99.9\%} & 0.5\% \\
``i will get plate 1'' & plate-1  & 99.0\% & \textbf{4.0\%} \\
\bottomrule
\end{tabular}
\end{table}

\vspace{-0.5\baselineskip}
\section{Related Work}
\label{sec:related}
\vspace{-0.5\baselineskip}
MAVIC relates to prior work on learning under multiple competing objectives. While typically studied in adversarial contexts, recent work in backdoor attacks uses reward constructions to induce agents to optimize temporally interleaved behaviors within a single policy \citep{rathbun_sleepernets_2024}. MAVIC is inspired by this perspective but differs fundamentally in objective: rather than embedding hidden behaviors, we address a structural inconsistency in value estimation arising from stochastic instruction switching, and correct it at the level of Bellman updates.

A large body of work studies language-guided decision making using VLMs and LLMs \citep{huang_language_2022, liu_interactive_2023, stone_open-world_2023, kim_openvla_2024}. These approaches often employ hierarchical decomposition, where high-level language reasoning generates subgoals executed by low-level controllers \citep{shi_hi_2025}. While effective for instruction following, such methods are computationally expensive and struggle with long-horizon coordination, particularly in multi-agent settings. In contrast, MAVIC operates at the control level, treating instructions as stochastic conditioning signals and avoiding expensive language inference at test time.

Outside of LLM-based approaches, prior work incorporates language, safety requirements or human desiderata into RL via reward shaping, temporal logic constraints, or policy conditioning \citep{shah_follownet_2018, rnn_ver, aydeniz2024aamas, liu_grounding_2023, holk_predilect_2024, eps_retrain, ecai_marl, shi_yell_2024, cui_no_2023}. These methods typically assume that standard Bellman updates remain valid when instructions modify rewards or objectives. However, as we show, this assumption breaks down when instructions arrive asynchronously and interrupt temporally extended actions, leading to cross-contamination in value estimates. MAVIC differs fundamentally by modifying the Bellman backup itself, rather than the reward or policy alone, to enforce consistency under stochastic instruction switching in multi-agent settings.

Standard multi-task and goal-conditioned RL methods learn policies across multiple tasks or goals, including partially observable multi-agent settings~\citep{omidshafiei_deep_2017,chen_variational_nodate}. These formulations generally treat the active task or goal as fixed over a rollout or training instance rather than as an objective that can change mid-trajectory and interrupt a temporally extended action. Our ``Switch'' baseline is a natural adaptation of this independent-context paradigm. MAVIC instead targets the Bellman-boundary inconsistency created by within-episode objective changes and can therefore be viewed as complementary to task/goal conditioning rather than a replacement for it.

%===============================================================================

\vspace{-0.5\baselineskip}
\section{Conclusion}
\label{sec:conclusion}
\vspace{-0.5\baselineskip}
We introduced MAVIC, a value correction method for following dynamic human instructions in asynchronous cooperative settings. We showed that naively conditioning rewards on instructions induces coupling between value estimates and proposed a value-corrected reward mechanism that prevents instruction-following returns from contaminating base task learning. Combined with an interruption-aware macro-action formulation, MAVIC enables consistent value estimation even under forced early termination. This yields a unified instruction-conditioned policy that generalizes across instruction phrasings via a frozen language encoder. Empirically, MAVIC achieves strong instruction compliance while preserving task performance across increasingly complex environments, outperforming MARL baselines. Additional appendix analyses show robust recovery under interruptions, value-estimation noise, and stochastic transitions.

These results highlight value-level decoupling as a fundamental mechanism for enabling MARL to robustly adapt to dynamic human input in real-world environments. Extending MAVIC to larger teams, non-predefined macro-actions, and open-ended instruction grounding remains future work.

%===============================================================================
\clearpage

\begin{ack}
The authors thank the reviewers for their insightful and constructive feedback, which has substantially
improved the quality of this work. We also thank everyone from PARCs lab for feedback and also Shuo Liu, Andrea Baisero, Rupali Bhati, Xueguang Lyu, and everyone from LLPR Lab for helpful discussions and feedback throughout the project. 

This work was supported in part by the Italian Ministry of University and Research (MUR) through the Italian Science Fund (FIS 3), project \textit{BeTrusty} (Grant FIS-2024-05614).
 
This work was also supported in part by Xiang Zhi Tan.

\end{ack}

%===============================================================================

% NeurIPS explicitly requires a style to be defined when using natbib 
% (which neurips_2026 loads by default). plainnat or unsrtnat are standard.
\bibliographystyle{plainnat} 
\bibliography{sections/references_v2, sections/other_references}

\clearpage
% \appendix
% \label{app:appendix}
% \input{sections/appendix}

\clearpage

\end{document}